\documentclass[runningheads]{llncs}
\usepackage{graphicx}
\usepackage{subfig}
\usepackage{booktabs}
\usepackage{multirow}
\usepackage{algorithm2e}
\usepackage{listings}
\usepackage{cite}
\usepackage{tablefootnote}
\begin{document}
\title{Cognitive system to achieve human-level accuracy in automated assignment of helpdesk email tickets}
\titlerunning{Automated Email Assignment}
\author{Atri Mandal\inst{1} \and Nikhil Malhotra\inst{2} \and
Shivali Agarwal\inst{1} \and Anupama Ray\inst{1} \and Giriprasad Sridhara\inst{1}}
\authorrunning{Mandal et al.}
\institute{IBM Research AI, Bengaluru, India \and
IBM Global Technology Services, Bengaluru, India
\email{\{atri.mandal,nikhimal,shivaaga,anupamar,girisrid\}@in.ibm.com}}
\maketitle              % typeset the header of the contribution
\vspace{-3mm}
\begin{abstract}
Ticket assignment/dispatch is a crucial part of service delivery business with lot of scope for automation and optimization. In this paper, we present an end-to-end automated helpdesk email ticket assignment system, which is also offered as a service. The objective of the system is to determine the nature of the problem mentioned in an incoming email ticket and then automatically dispatch it to an appropriate resolver group (or team) for resolution. \par
The proposed system uses an ensemble classifier augmented with a configurable rule engine. While design of classifier that is accurate is one of the main challenges, we also need to address the need of designing a system that is robust and adaptive to changing business needs. We discuss some of the main design challenges associated with email ticket assignment automation and how we solve them. The design decisions for our system are driven by high accuracy, coverage, business continuity, scalability and optimal usage of computational resources. \par
Our system has been deployed in production of three major service providers and currently assigning over 40,000 emails per month, on an average, with an accuracy close to 90\% and covering at least 90\% of email tickets. This translates to achieving human-level accuracy and results in a net saving of about 23000 man-hours of effort per annum.

\keywords{Cognitive email assignment \and Helpdesk automation \and Ticket resolver group \and Smart dispatch \and Ensemble classifiers}
\end{abstract}
\section{Introduction}
The landscape of modern IT service delivery is changing with increased focus on automation and optimization. Most IT vendors today, have service platforms aimed towards end-to-end automation for carrying out mundane, repetitive labor-intensive tasks and even for tasks requiring human cognizance. One such task is ticket assignment/dispatch where the service requests submitted by the end-users to the vendor in the form of tickets are reviewed by a centralized dispatch team and assigned to the appropriate service team i.e. resolver group.\par
%For systems in production, end-users submit service requests to the vendor in the form of tickets that represent a specific computer related problem experienced by the end user (e.g., a failed printing job, user authentication expiry, data backup etc.). The tickets are usually assigned to an engineer in a two-step process. First, incoming tickets are received by a member of a centralized dispatch team who reviews the problem description text to understand which service team (resolver group) is responsible for addressing the ticket, and then dispatches it to that group of engineers. Next, within the group, the assignment of the ticket to a specific engineer may be done by the group lead, or an engineer may volunteer for the same. \par
The dispatch of a ticket to the correct group of practitioners is a critical step in the speedy resolution of a ticket. Incorrect dispatch decisions can significantly increase the total turnaround time for ticket resolution, as observed in a study of an actual production system \cite{AgarwalKdd}. %When such delays occur, it causes customer dissatisfaction as well as monetary penalties for the vendor due to Service Level Agreement (SLA) breaches. 
Several factors make the dispatcher's job challenging such as requirement of knowledge of the IT portfolio being managed, roles and responsibilities of the individual groups, ability to quickly parse the ticket text describing the problem and map it to the right group, which is often not straightforward given the heterogeneous and informal nature of the problem description. 
A number of different approaches have been proposed for automating ticket dispatch \cite{AgarwalKdd}\cite{ShaoVldb}\cite{ShaoKdd}\cite{Parvin}. 
Although automated email assignment may look like a simple text classification problem at first glance it becomes quite complex and challenging when considered at industry scale. 
%Thirdly the problems assigned to resolver groups themselves slowly change over time. This may not happen over a week or month but over several quarters. As such a once-trained model becomes outdated over time as it cannot effectively assign tickets mentioning new problems or old problems with a different terminology. \par
\vspace{-5mm}
\subsection{Main Contributions}
In this paper we present a readily deployable end-to-end automatic email dispatch system, which has the following key features:
% * <sgiriprasad@gmail.com> 2018-06-05T08:32:27.414Z:
% 
% >  
% need sub-sections ? without it , we can save space
% 
% ^.
\begin{enumerate}
  \item An ensemble based classification engine that uses supervised machine learning to understand the nature of the problem from free unstructured email text and assign accurately. The choice of ensemble is based on the results of comprehensive study performed with various machine learning and deep learning models as presented in section~\ref{ssec:classificationmodels}.   
  \item A rule engine with a customer-independent framework for rule specification to ensure business continuity and handle domain specific content missed by the ensemble classifier.
\end{enumerate}
\textit{We present a comprehensive study of the efficacy of different machine-learning and deep-learning algorithms in helpdesk email ticket classification}. The results are presented with real customer data from three different datasets – with the largest of them having more than 700,000 emails and as many as 428 resolver groups. We were able to achieve human level accuracy with more than 90\% coverage on all the datasets with the proposed system using minimal computational resources. % that uses an ensemble of traditional machine learning algorithms and requires  
\par
\textit{To the best of our knowledge, this is the first time that human-level accuracy has been reported for an assignment engine at this scale of automation, delivered consistently across datasets of varying size.} 
The remainder of the paper is organized as follows. Section \ref{sec:relatedwork} describes the related work. Section \ref{sec:systemoverview} gives an overview of the system used for ticket classification and section \ref{sec:assignmentcomponents} discusses the different components of the system. In section \ref{sec:results} we present our experimental results while we conclude in section \ref{sec:conclusion}.
\vspace{-2mm}
\section{Related Work}
\label{sec:relatedwork}
Ticket dispatch has been addressed by \cite{AgarwalKdd} using Support Vector Machines and discriminative keyword approach. They propose semi-automated approach based on confidence scores. We have surpassed their work to i) reach human level accuracy using advanced ensemble techniques for automated dispatch, ii) scale it to hundreds of resolver groups and iii) incorporate retraining strategies to adapt to changing data. Several other researchers have studied different aspects of the problem of routing tickets to resolver groups \cite{ShaoVldb}\cite{ShaoKdd}\cite{Parvin}. The work in \cite{ShaoKdd} approaches the problem by mining resolution sequence data and does not access ticket description at all. Its objective is to come up with ticket transfer recommendations given the initial assignment information. The work in \cite{ShaoVldb} mines historical ticket data and develops a probabilistic model of an enterprise social network that represents the functional relationships among various expert groups in ticket routing. Based on this network, the system then provides routing recommendations to new tickets. This work also focuses on ticket transfers between groups (given an initial assignment) without looking at the ticket text content. The work in \cite{Parvin} is different and approaches the problem from a queue perspective. This is more related to the issue of service times and becomes particularly relevant when the ticket that has been dispatched to a group needs to be assigned to an agent. There are some papers, which apply text classification techniques to handle tickets~\cite{GargiIcsoc}\cite{ZengTnsm}. The idea is that once ticket category is identified, then the assignment to resolver groups can be done by manual dispatchers quickly. However, none of the works talk about the scale and retraining required in real-life deployment. In \cite{LuccaIcsm} tickets are automatically classified based on description to route them to the right group. However, the work was applied on a small ticket set with only 8 groups. The work in \cite{KadarSRII} attempts to classify the incoming change requests into one of the fine-grained activities in a catalog. Some other works \cite{Potharaju} and \cite{AgarwalBook} talk about a holistic approach of ticket category classification, cause analysis and resolution recommendation. However, they do not automate the process of assignment.

\section{System Overview}
\label{sec:systemoverview}
\vspace{-3mm}
\begin{figure}
\centering
\includegraphics[width= 10cm, height= 5cm]{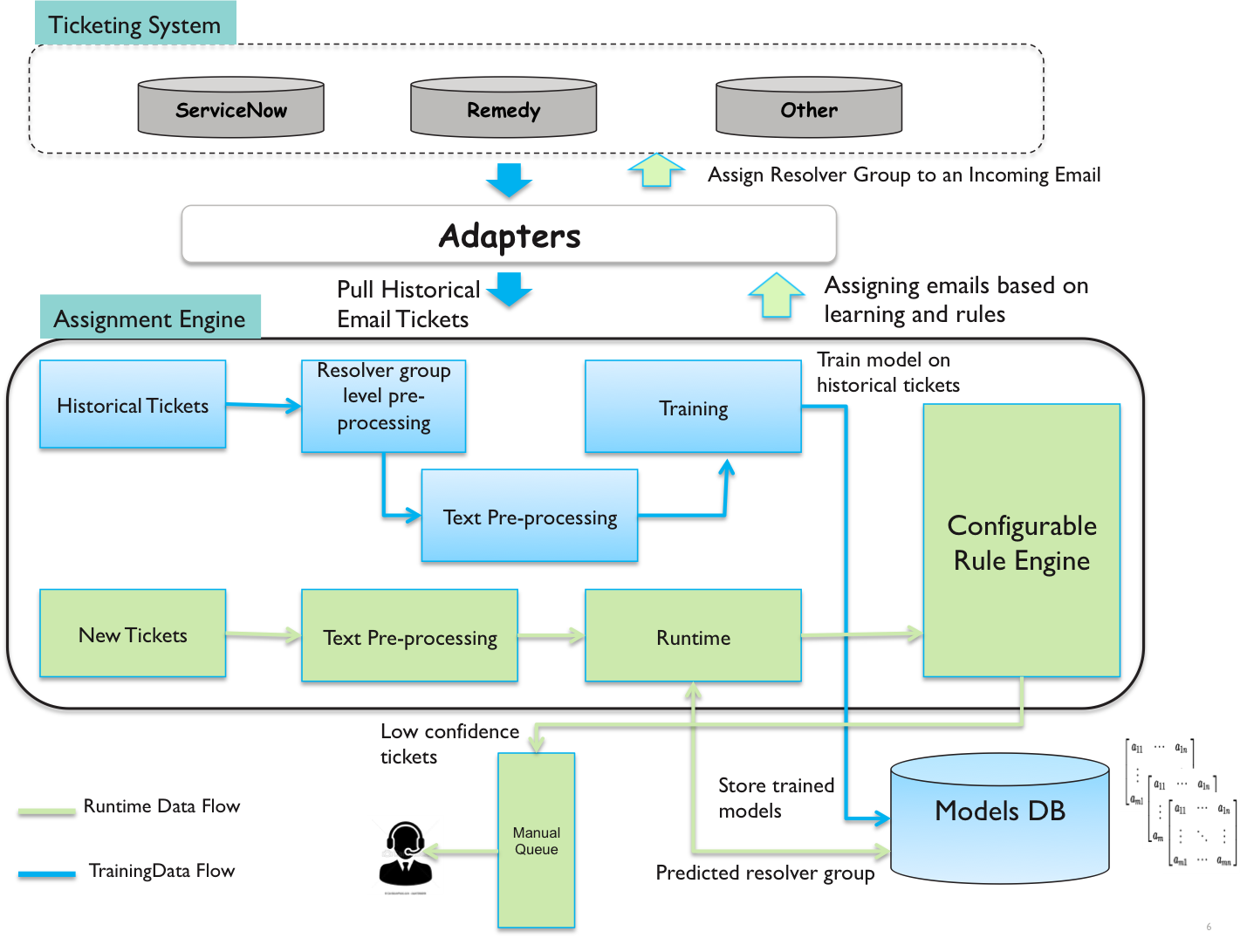}
  \caption{Architecture of the proposed system}\label{fig:arch}
 \end{figure} 
Figure \ref{fig:arch} shows the system architecture along with the data flow diagram.  Historical email ticket data is downloaded from the ticketing tool (e.g. Remedy or ServiceNow) using custom-built adapters. The downloaded emails are passed through two stages of pre-processing for data enrichment. The resolver group level pre-processing module uses techniques like resolver group merging, long tail cutoff etc. to reduce the noise in the email data. The training data is further enriched using text pre-processing methods. The enriched email data is then trained using an ensemble of machine learning classifiers and the trained models are stored in a database. \par
When a user sends an email to the helpdesk account a ticket is automatically generated and stored in the backend ticketing tool.  The newly generated tickets are downloaded by the adapter and classified using the runtime that consists of  ensemble classifier and the rule engine. The classification system returns a resolver group along with a confidence score. If the confidence score is above a configured threshold the ticket is routed to the returned resolver group. Otherwise the ticket is assigned back to manual queue for inspection by human agent. The combination of ensemble classifier and rule engine ensures that a high percentage of tickets (more than 90\%) are classified automatically by our system with a low error rate.  

% * <sgiriprasad@gmail.com> 2018-06-05T08:33:22.762Z:
% 
% > Definition of some key terms: 
% seems unnecessarily complex especially the math formulas
% 
% ^.
\section{Assignment Engine Components}
\label{sec:assignmentcomponents}
\subsection{Preparation of Training Data}
\label{ssec:preparation}
Most large companies nowadays use ticketing tools like Remedy or ServiceNow to maintain tickets obtained from various channels (voice, email, web etc.) by the helpdesk. The ticketing tool organizes the email data into structured fields containing relevant information about the ticket e.g. incident number, incident type, date of creation, description, assigned group etc. 
We use custom adapters to connect to the ticketing tool and extract fields relevant for training. Currently the adapter extracts only the text portion of the email (viz. email subject and body) along with the resolver group for training.  The data collected by the adapter is then converted into a format readable by the classifier. The steps involved in training data preparation are described below.  
\vspace{-3mm}
\subsubsection{Resolver group level pre-processing}
This type of pre-processing is a one-time effort required during customer on-boarding phase.  The purpose of this pre-processing is to reduce noise in the training data.  We reduce noise and enrich training data for the resolver groups using the following techniques:

\subsubsection{Merging related resolver groups}
Some of the resolver group labels in the training data can be merged. Merging increases the size of the training data and at the same time reduces the number of unique labels thus improving training accuracy. We found that there are at least two types of resolver groups that can be merged for assignment purpose.
\textbf{a) Resolver groups with varying escalation levels}: Firstly there are some tickets for which the final assigned group is one of various escalation levels (e.g. Tier1, Tier2 etc.) of the same resolver group.  However the helpdesk often assigns directly to only one of these escalation levels. Escalations to higher levels usually happen after additional information is sought from the customer. But it is enough to assign the ticket initially to the default escalation level. 
\textbf{b) Zone specific resolver groups}: Sometimes tickets are meant to be resolved based on the user zone or location. In that case the tickets are assigned to a particular zone and is resolved by an agent locally. To merge these resolver groups we create a new label and assign all the zone specific tickets to this label. After the initial assignment, the correct resolver group can be inferred based on some other fields in the ticket e.g. end-user location or originator email-id etc. using the rule engine.

\subsubsection{Long tail cutoff}
\label{ltc}
%We study the impact of the count of resolver groups on the accuracy of training data. 
We observed that in most of the datasets there are a large number of resolver groups with very few samples.  If we plot a histogram of frequencies these groups will constitute more than 80\% of the resolver groups but less than 5\% of training data. Our studies indicate that, if the long tail is included in training, the overall accuracy of classification goes down along with a significant increase in training time and model size. By restricting the number of resolver groups in training we reduce noise significantly and also avoid class imbalance, hence the increase in accuracy.  Additionally, the resolver groups, which fall in the long tail, can often be predicted better using the rule engine and using some augmentation techniques. 
As such our strategy was to divide the downloaded historical data into 2 parts viz. 
		I\textsubscript{H} = I\textsubscript{T} + I\textsubscript{L} 
where I\textsubscript{H} is the complete data downloaded for training, I\textsubscript{T} is the data used for training classifiers and I\textsubscript{L} is the long tail. Resolver groups belonging to I\textsubscript{T} will be classified using trained models while those belonging to I\textsubscript{L} will be handled exclusively by the rule engine. In our system we use the above strategy to retain at least 98\% of data while cutting down the resolver group count to less than 20\%.
% To ensure better training we augmented the training data using the following methods: 
% \begin{itemize}
% \item Direct addition of key terms or phrases – In this method we modify the existing sample by adding some key terms from helpdesk support guidelines. The support guideline refers to the instruction manual provided to the helpdesk agents for manual assignment. If such guidelines are not available we have to manually obtain the key terms by interacting with helpdesk agents. The key phrase or term is added to the beginning or end of the email conversation.
% \item Addition as skip-grams – In this method the key phrase is split up into individual words and is added as skip-grams to the email data. 
% \item Duplication of samples:
% For some datasets the helpdesk manual may not be available. In such cases we simply duplicate a sample from the sparse resolver group without any modifications. The duplication is done multiple times till the sample count reaches $S\textsubscript{min}$. 
% \end{itemize}
\vspace{-2mm}
\subsection{Classification Models}
\label{ssec:classificationmodels}
This section presents our study on the performance of various machine-learning classifiers in classification of email data, in terms of accuracy and training time, although the training is offline. For training the classification models, we concatenate the subject and the body of the email(description) with a space in between and use the resulting string as our training data. The resolver group acts as the label for our training data. Table~\ref{tab:MLmodels} and Table~\ref{tab:NN} show the impact of various traditional machine learning models \cite{Mitchell} and deep neural network models that were used. In order to improve accuracy and coverage of the overall service, we use an ensemble \cite{kuncheva2004combining}. Each pair of models were combined, and the final ensemble classifier was chosen based on the accuracy and coverage. As explained in section \ref{ltc}, rule engine is important to handle the long tail in class distribution and the final chosen ensemble classifier in combination with the rule engine forms the classification module of the service. 
\vspace{2mm}
\begin{table*}
\centering
  \caption{Comparison of various Machine Learning Algorithms w.r.t. Accuracy and Training Time}
  \label{tab:MLmodels}
  \begin{tabular}{|cl|c|c|c|c|c|c|c|}
    \toprule
    %\multirow{2}{*}{}& & LinearSVC & KNN & LR& m-NB & RF & Adaboost & Gradient Boosting\\&\\
    \multirow{2}{*}{}& & LinearSVM & KNN & LR& m-NB & RF & Adaboost & Gradient\\&{}&{}&{}&{}&{}&{}&{}& Boosting\\
    %multicolumn{2}{}{}& & LinearSVC & KNN & LR& m-NB & RF & Adaboost & \multirow{2}{Gradient}{}& \multirow{2}{}{Boosting}\\
    \hline
    \multirow{2}{*}{Dataset A} & Accuracy(\%)& 87.3 & 80.12 &79.48 & 72.68 & 81.41 & 31.5 & 75.6 \\ & Train-time(s) & 7.8 & 260.5 & 43 & 17.3 & 363.75 & 4561 & 8612\\
%\cmidrule(lr){2-8}
\hline
    \multirow{2}{*}{Dataset B} & Accuracy(\%)& 83.42 & 72.58 & 79.95 & 64.19 &74.91 & 32.98 & 65.1\\& Train-time(s) & 76.12 & 2218.65 & 404.05 & 22.18 & 7190.16 & 332.97 & 95320.1\\
    \hline
    \multirow{2}{*}{Dataset C} & Accuracy(\%) & 86.339 & 67.57 & 84.29 &63.97 & 76.99 & 30.43 & 61.47\\ & Train-time(s) & 1001.06 & 1921.7 &2992 & 167.5 & 20799.6 & 1288.63 & 126960\\
 \bottomrule
\end{tabular}
\end{table*}
\vspace{-3mm}
\begin{table*}
\centering
  \caption{Comparison of various Deep neural networks w.r.t. Accuracy and Training Time}
  \label{tab:NN}
  \begin{tabular}{|cl|c|c|c|c|c|c|}
    \toprule
    \multirow{2}{*}{}& &  MLP & CNN-WE& LSTM-WE & CNN-G & LSTM-G & CNN-LSTM-G\\& \\
    \hline
    \multirow{2}{*}{Dataset A} & Accuracy(\%) & 85.8 & 74 & 76.94 &74.01 & 71.64 & 73.24\\ & Train-time(s) & 184.12 & 183.75 & 5546.6 & 160.56 & 9833.7 & 1844.8 \\
%\cmidrule(lr){2-8}
    \hline
    \multirow{2}{*}{Dataset B} & Accuracy(\%) & 80.87 & 77.75 & 79.35 & 76.23 & 80.37 & 77.7\\
& Train-time(s) & 10858.15 & 8680.35 & 86651.57 & 1926 & 89280.94 & 23229.47\\
    \hline
    \multirow{2}{*}{Dataset C} & Accuracy(\%) & 83.3 & 78.8 & 78.14 & 79.1 & 83.51 & 81.33\\
& Train-time(s) & 2779 & 4000.9 & 90149.9 & 9522.12 & 687483 & 116583.22\\
     \bottomrule
\end{tabular}
\end{table*}
\vspace{-4mm}
\subsubsection{Training the classifiers}
We convert the training data samples into word vector representation before applying machine-learning algorithms. We observed that using tf-idf representation increased the accuracy of traditional machine learning algorithms for all datasets by at least 3-4\%. Another observation was that using bigrams also improved the accuracy for some datasets. Intuitively we can argue that this is so because some bigrams like ‘account creation’, ‘account deletion’, ‘password reset’ etc. are useful indicators in deciding the resolver group.  The hyperparameters were chosen experimentally over 10-fold cross-validation on the datasets.\par  %The results are shown in the next section. 
However, for learning deep neural networks, tf-idf representation being extremely sparse is not useful. Distributed representation of text creates a dense, low-dimensional representation and is perhaps the main reason why deep learning saw major breakthroughs in natural language processing \cite{Mikolov:2013:DRW:2999792.2999959}. There are primarily two methods of learning the word embeddings: one in which word embeddings are learnt while training the neural network; and second using pretrained word vectors. We experimented with both methods for classification (models learning word embeddings being referred to CNN-WE, LSTM-WE, and CNN-LSTM-WE in Table \ref{tab:NN}), and pretrained word-vector representations (100-d GloVe vectors) \cite{pennington2014glove} referred to as CNN-G, LSTM-G and CNN-LSTM-G.  
\subsection{Rule Engine}
\label{ssec:ruleengine}
The rule engine is one of the key components of our end-to-end system and is used to capture domain specific elements in training data which machine-learning or deep-learning classifiers are not able to detect as given below: 
\newline
\textbf{Resolver group perturbations driven by business decisions:}
Often resolver groups are either renamed or split or merged to form new resolver groups. These decisions are mostly taken to remove duplication of effort, or to address macro-economic changes. As most of these decisions are sudden, machine-learning models are not able to handle classification for the newly formed classes, which affects services in production. 
\newline
\textbf{Resolver groups belonging to the long tail:}
As discussed in section \ref{ssec:preparation}, 20\% of the classes account for 98\% of the tickets. The learning models are not trained on the remaining 80\% classes to reduce noise and avoid class imbalance. Although this improves classification accuracy and time, the model will never learn to predict these classes. For all these classes the rule engine is essential.
\newline
\textbf{Presence of resolver groups with similar or overlapping email format: }
Many helpdesk organizations use fixed templates for submission of certain types of issues. The same template can be used for multiple resolver groups. When these tickets are used to train the machine learning model, it learns the template structure rather than the actual content. So the classification accuracy is very low for such resolver groups. Rule-engine addresses this issue for the confusing classes to override the decision of the machine learning classifier.
\par
The rule engine is designed to have a customer independent framework for rule specification, which is easy to configure (using regular expressions and rules specified using values of certain ticket parameters e.g. email subject, description etc. and the output of the ensemble classifier). The rule engine can override the output of the ensemble classifier in certain cases. 

\subsection{Email Ticket Dispatcher}
The email ticket dispatcher actually assigns the ticket to a specific resolver group and updates the ticket. The dispatcher combines the results of the two classifiers and rule engine using a dispatch algorithm  to output the final prediction and confidence score. If the confidence score of the final result is below the configured threshold the ticket is assigned to the manual queue. 

% \begin{lstlisting}
% result = [None, 0.0]
% ensemble_cutoff = min (MLC1_cutoff, MLC2_cutoff)
% if  confidence(MLC1) >= MLC1_cutoff  then:
% 	result  = [resolver_group(MLC1), confidence(MLC1)]
% else if  confidence(MLC2) >= MLC2_cutoff then:
% 	result  = [resolver_group(MLC2), confidence(MLC2)]
% else  if  resolver_group(MLC1) == resolver_group(MLC2) then:
% 	result  = [resolver_group(MLC1),  ensemble_cutoff]
% #next invoke the rule engine, irrespective of the previous outcome
% result =  RuleEngine([F1, F2, F3, result])
% if result.resolver_group <> None:
% 	return result
% else:
% 	assign ticket to manual queue 
% \end{lstlisting}

\vspace{-7mm}
\begin{table}
\centering
\small
\caption{Dataset details and results}
  \label{tab:dataset}
  \begin{tabular}{|c|c|c|c|}
    \hline
   &  \textbf{Dataset A} & \textbf{Dataset B} & \textbf{Dataset C} \\
    \hline
    \textbf{Number of email tickets(N)} & 11562 & 423343 & 712320\\
    \hline
    \textbf{Number of resolver groups} & 70 & 403 & 428\\
\hline    
    \textbf{Duration of the dataset} & 6 months & 12 months & 15months\\
  \hline     
    \textbf{Ensemble Accuracy(X\textsubscript{acc})} & 90.07\% & 86.17\% & 89.61\%\\
    \hline
    \textbf{Ensemble Coverage(X\textsubscript{cov})} & 93.67\% & 92.88\% & 93.83\%\\
\hline    
    \textbf{Assignment Engine Accuracy(E\textsubscript{acc})} & 92.73\% & 88.66\% & 92.13\%\\
  \hline     
  \textbf{Assignment Engine Coverage(E\textsubscript{cov})} & 97.84\% & 93.3\% & 95.5\%\\
  \hline
    \end{tabular}
\end{table}
\vspace{-7mm}
\section{Experimental Results}
\label{sec:results}
This section enumerates the results of the experimental setup of the assignment engine. For evaluation we have used real datasets from three major helpdesk service providers. The datasets are from two different domains viz. telecom and supply-chain/logistics. To preserve client confidentiality we henceforth refer to these datasets as Dataset A, Dataset B and Dataset C respectively. The datasets were divided into training and test sets with a 90:10 split and we used 10-fold cross-validation on the datasets. All our experiments were run on a  NVIDIA  Tesla  K80  GPU cluster with 4 CUDA-enabled nodes. We use open source machine-learning libraries viz. Python scikit-learn and Keras for our experiments. %The deployed system is similar to our experimental setup but not identical. The numbers in production may vary slightly. For confidentiality reasons we cannot reveal exact details of the production setup and accuracy results. 
The dataset statistics as well as the final accuracy numbers achieved by our system are described in Table \ref{tab:dataset}. Please note that the deployment setup is similar to our experimental setup but not identical; so numbers in production may vary slightly. The details about production setup and results are not included to preserve confidentiality.

\begin{figure}
\centering
\subfloat[]{\includegraphics[width=6cm, height=3cm]{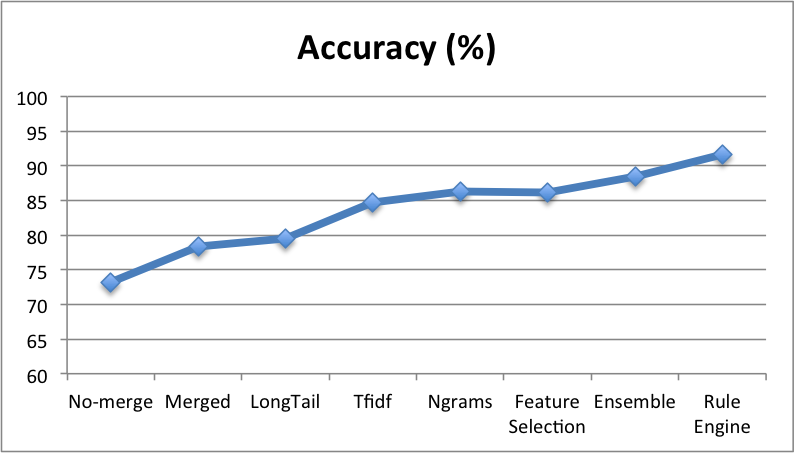}}
\subfloat[]{\includegraphics[width=6cm, height=3cm]{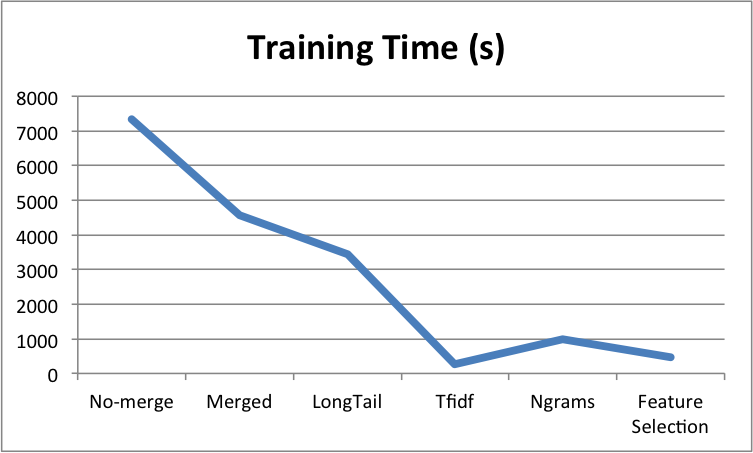}}
\caption{Effect of different optimization techniques (a) Accuracy trend (b) Training time.}
\label{fig:acc_time}
\end{figure}
\subsection{Accuracy and Training Time}
Figure \ref{fig:acc_time} shows how the training techniques and preprocessing affect the accuracy of prediction and training time. It shows the gradual increase in accuracy and corresponding decrease in training time as we apply each technique(shown in X-axis) incrementally. The accuracy and training time charts are shown for only one of the datasets viz. dataset C which is our largest dataset – but the trend is fairly similar across other datasets as well. 
\subsection{Human Accuracy vs. Assignment Engine Accuracy }
We next look at the optimal selection of algorithms that maximize accuracy and coverage. We assert that for business purposes the algorithms need to have at least human-level accuracy with high enough coverage. \par
\begin{figure}
\centering
\subfloat[]{\includegraphics[width=6cm, height=3.5cm]{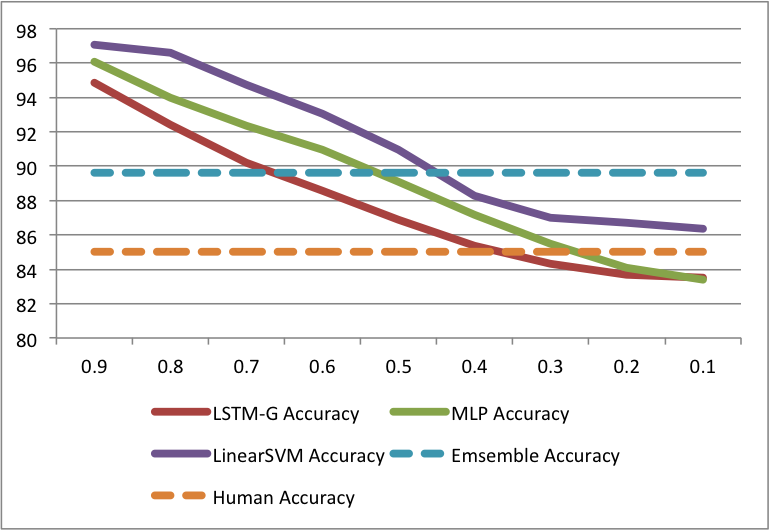}}
\subfloat[]{\includegraphics[width=6cm, height=3.5cm]{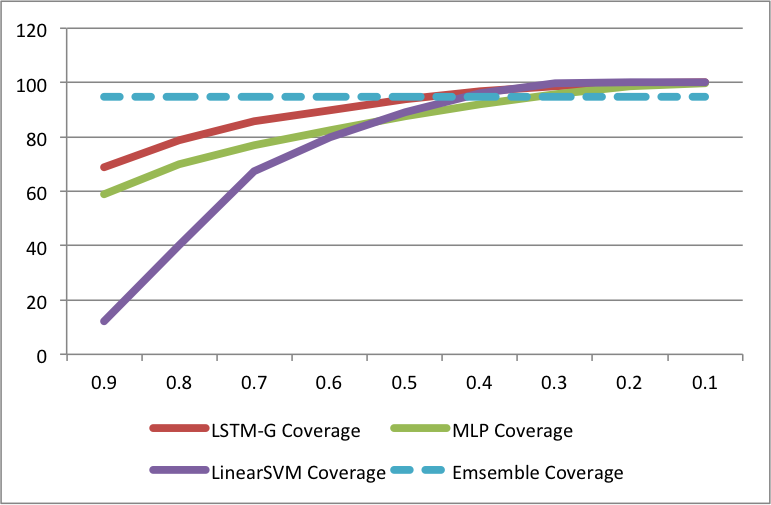}}
\caption{At different confidence thresholds(a) Assignment accuracy (b) Assignment coverage.}
\label{fig:acc_cov}
\end{figure}

 
To compute human accuracy we mined audit logs of the ticketing systems. Our experiments reveal that across all datasets the accuracy achieved by human agents is about 85\%. Therefore we select the confidence threshold  such that the expected accuracy of prediction is at least 85\%. Figure \ref{fig:acc_cov} show the performance of the best three algorithms at different confidence levels (ranging from 0.1 to 0.9). For dataset C a combination of linear SVM (confidence$\geq$ 0.5) and MLP (confidence$\geq$ 0.6) gave a slightly higher accuracy(89.61\%) than that of LSTM-G(confidence$\geq$0.5) and linear SVM(X\textsubscript{acc}=88.38\%), although the individual accuracy was marginally higher for LSTM-G compared to MLP. For this reason, as also for other practical considerations like memory and CPU constraints as well as training time our deployment in production uses an ensemble of linear SVM and MLP. For the other two datasets SVM and MLP were clear winners.  

%\subsection{Estimation of cost saving}
%Here we give an estimate of cost saving based on our experimental results in Table 3. Let's assume that for dataset B (the largest dataset) the total number of incoming tickets per month is 40,000. The assignment engine has a coverage of 95.5\% for this dataset which translates to about 38200 tickets. Assuming a human agent takes about 3 min to read and assign each ticket this results in a total saving of: $38200 \times 3 = 114600$ mins $= 1910 $ hours per month. As such the total saving of human effort per annum will be: $1910 \times 12 = 22920$ hours(approx). 
%\vspace{-7mm}
\subsection{Observations}
There are three main takeaways from our experimental results above.
The most important observation is that our assignment engine performs better (both in terms of accuracy and coverage) than all traditional machine-learning and deep learning algorithms.
Secondly we can see that simple machine learning algorithms like SVM and MLP are often better than more computationally expensive deep learning algorithms in the task of helpdesk email assignment automation. This result is somewhat surprising and unexpected, but is very significant from a product development standpoint as these algorithms are easy to implement, require minimal computational resources and provide better performance at runtime. \par
Another very important observation is that LSTM accuracy increases with the size of the dataset and for the largest dataset (dataset C) it outperforms MLP. However it must be noted that LSTM is computationally very expensive and may not be feasible if there are infrastructure constraints like memory, storage and cluster size. Thus our results indicate that an ensemble of SVM and MLP will be a good trade-off for most practical purposes but if we have a large enough dataset and infrastructure is not a concern then the best choices are SVM and LSTM-glove. \par

\section{Conclusion and Future Work}
\label{sec:conclusion}
%We gathered email based problem descriptions from three different IBM customers, totaling over 700000 emails distributed across more than 500 resolver groups. 
We have proposed email ticket assignment engine that uses an ensemble of machine learning techniques to perform automated dispatch. We combined our ensemble classifier with a configurable rule engine to detect domain specific utterances for optimal performance. Our system achieves human-level accuracy and has already been deployed for three customers in production. 
%We looked at different strategies for processing the raw email data and also how to enhance the training data using augmentation techniques, resolver group merging, long tail cutoff etc. 
%We also looked at how to optimally select the training set and window, so that higher priority is given to recent data and at the same time training accuracy is not impacted.
%We compared different machine learning and deep learning methods for training classifiers on the email data. And finally we looked at how to combine the best machine learning algorithms into an ensemble to achieve optimal accuracy and coverage.  
However, there are still some areas in the system like rule engine which need human intervention and can be automated. %Firstly the configurable rule engine requires helpdesk personnel to continuously configure rules for assignment based on keywords and phrases. This requires a lot of manual effort as rules can change over time and if not monitored continuously will lead to wrong assignment. 
In future, we want to solve the problem of automatically extracting rules based on data from misclassified emails. We would also like to handle concept drift in utterances for better retraining. Last but not the least, we need to enhance our assignment algorithm to handle email attachments.
%Secondly to maintain a good training set we currently require newly resolved tickets to be manually inspected for possible errors in human assignment. This prevents wrongly labeled data to be used for training. We can remove this requirement by mining audit logs to detect errors in assignment wherever possible. 
%Thirdly we need to look at email attachments in addition to text for classification. In a lot of cases users will only send a screenshot or log snippet to the code with hardly any text conversation. These cases cannot be handled by the present system as it relies solely on the text to understand the nature of the problem.  
%Last but not the least we need to handle concept drift. This is currently handled to some extent by using a sliding window of recent data as well as use of the rule engine. However in datasets where the impact of concept drift is high this method may not give good results over the long run. We need to come up with an effective active learning strategy to handle such scenarios. 

%\section{References}
\vspace{-3mm}
\bibliographystyle{splncs04}
\bibliography{cae_references}
\end{document}